\documentclass{llncs}
\usepackage[american]{babel}
\usepackage[T1]{fontenc}
\usepackage{times}
\usepackage{graphicx}
\usepackage{pdfpages}
\usepackage[all]{nowidow}
\usepackage{multirow}

\begin{document}
	
{\let\thefootnote\relax\footnotetext{Copyright \textcopyright\ 2020 for this paper by its authors. Use permitted under Creative Commons License Attribution 4.0 International (CC BY 4.0). CLEF 2020, 22-25 September 2020, Thessaloniki, Greece.}}

\title{Fake News Spreader Detection on Twitter\\ using Character $N$-Grams}
%%% Please do not remove the subtitle.
\subtitle{Notebook for PAN at CLEF \the\year}

\author{Inna Vogel \and Meghana Meghana}
\institute{Fraunhofer Institute for Secure Information Technology SIT\\ Rheinstrasse 75, 64295 Darmstadt, Germany\\
	\{Inna.Vogel, Meghana.Meghana\}@SIT.Fraunhofer.de}

\maketitle

\begin{abstract}
The authors of fake news often use facts from verified news sources and mix them with misinformation to create confusion and provoke unrest among the readers. The spread of fake news can thereby have serious implications on our society. They can sway political elections, push down the stock price or crush reputations of corporations or public figures. Several websites have taken on the mission of checking rumors and allegations, but are often not fast enough to check the content of all the news being disseminated. Especially social media websites have offered an easy platform for the fast propagation of information. Towards limiting fake news from being propagated among social media users, the task of this year's PAN 2020 challenge lays the focus on the fake news spreaders. The aim of the task is to determine whether it is possible to discriminate authors that have shared fake news in the past from those that have never done it. In this notebook, we describe our profiling system for the fake news detection task on Twitter. For this, we conduct different feature extraction techniques and learning experiments from a multilingual perspective, namely English and Spanish. Our final submitted systems use character $n$-grams as features in combination with a linear SVM for English and Logistic Regression for the Spanish language. Our submitted models achieve an overall accuracy of 73\% and 79\% on the English and Spanish official test set, respectively. Our experiments show that it is difficult to differentiate solidly fake news spreaders on Twitter from users who share credible information leaving room for further investigations. Our model ranked 3rd out of 72 competitors. 
\\

\end{abstract}

%%%%%%%%%%%%%%%%%%%%%%%%%%%%%%%%%%%%%%%%%%%%%%%%%%%%%%%%%%%%%%%%%%%%%%%%%%
\begingroup
\let\clearpage\relax
\include{texfiles/Introduction} %1
\include{texfiles/Related_Work} %2
\section{Dataset and Corpus Analysis}
\label{Dataset}

\newcommand{\trueuser}{$\mathcal{U}_{\emph{True }}$}
\newcommand{\fakeuser}{$\mathcal{U}_{\emph{Fake }}$}

\noindent To train our system, we used the PAN 2020 author profiling corpus\footnote{https://zenodo.org/record/3692319\#.XrlnomgzZaQ} proposed by Rangel et al. \cite{PAN2020_Data_Rangel}. The corpus consists of 300 English (EN) and Spanish (ES) Twitter user accounts each. The tweets of every Twitter user are stored in an XML file containing 100 tweets per author. Every tweet is stored in a \textit{<document>} XML tag. The tweets were manually collected and fact-checked. The dataset is balanced which means the data refers to an equal distribution of class instances. Half of the documents per language folder are authors that have been identified sharing fake news. The other half are texts from credible users. Table \ref{tab:excerpt} shows excerpts from the data. Every author received an alphanumeric author-ID which is stored in a separate text file together with the corresponding class affiliation. 
For training and testing, we split the data in the ratio 70/30. The gold-standard can only be accessed through the TIRA \cite{tira2019} evaluation platform provided by the PAN organizers. The results are hidden for the participants.    

\begin{table*}[h!]
	\centering
	\caption{English (EN) and Spanish (ES) excerpts from the PAN 2020 Twitter ``Fake News Spreader'' data.}	
	\label{tab:excerpt}
	
	%\resizebox{\textwidth}{!}{%
	\begin{tabular}{p{6cm}p{6cm}}
		\hline
		\multicolumn{1}{c}{\textbf{EN and ES True News Tweets}} & \multicolumn{1}{c}{\textbf{EN and ES Fake News Tweets}} \\ \hline
		``RT \#USER\#: Best dunk of the contest no doubt about it. Aaron Gordon robbed again \#URL\#'' & ``Jay-Z Must Give Beyonce \$5 Million Per Child They Have Together Due to Crazy Prenup…\#URL\#'' \\
		``RT \#USER\#: Sure would be an interesting day to read a book that examines Trump’s obsession with the king-like powers of his offic…'' & ``RT \#USER\# \#USER\# When Obama was tapping my phones in October, just prior to Election!'' \\
		``A Data-Driven Approach Aims to Help Cities Recover After Earthquakes \#URL\#'' & ``Why Trump lies, and why you should care - The Boston Globe \#URL\#'' \\ \hline
		
		``Javier Cámara ya es el líder más valorado de los españoles por delante de Pedro Sánchez, según una encuesta \#URL\# \#URL\#'' & ``Dictadura pura y dura toma tasas y todos felicices \#URL\#'' \\
		``Me gusta la foto. Una foto con variedad, diversidad. Me da la impresion que con más sonrisas que otras. \#URL\#'' & ``GANAR DINERO AHORA ES FACIL – Google te paga 15 dólares por contestar encuestas \#URL\# \#URL\#'' \\
		``Navidad en RD: son 3 días gozando, luego 362 llorando y deseando mal a los demás. Dejen su hipocresía !!'' & ``Ortega Smith: `VOX expulsará de España a todos los inmigrantes ilegales' \#URL\#'' \\ \hline
	\end{tabular}
	%}
\end{table*}

\noindent As can be seen in Table \ref{tab:excerpt}, the Twitter specific tokens hashtags, URLs and user mentions were replaced by the providers with the following placeholders: \emph{\#HASHTAG\#}, \emph{\#URL\#} and \emph{\#USER\#}. Prior to the feature engineering, we analyzed the distribution of different tokens. Additionally, we determined the sentiment of each tweet (positive, negative, or neutral) using \emph{TextBlob}\footnote{https://textblob.readthedocs.io/en/dev}. For recognizing the named entities (NER), we used the Python library \emph{spaCy}. Table \ref{tab:feature-dist} shows some key insights for both languages. \\

\begin{table}[htbp]
	\centering
	\caption{Feature distribution of the fake news (Fake) and true news (True) spreaders}
	\label{tab:feature-dist}	
	\begin{tabular}{p{4cm}p{2cm}p{2cm}p{2cm}p{2cm}}
		\hline
		\multicolumn{1}{c}{}    & \multicolumn{2}{c}{\textbf{English}}  & \multicolumn{2}{c}{\textbf{Spanish}}  \\ \cline{2-5} 
		\textbf{Features}       & \textbf{True} & \textbf{Fake} & \textbf{True} & \textbf{Fake} \\ \hline
		Unique Tokens             & 24,050    & 23,809    & 32,802    & 27,932    \\ \hline 
		Emojis Total             & 1,614  & 522    & 3,867  & 1,629  \\
		Emojis Unique            & 325    & 145    & 603    & 301    \\ \hline
		Neutral Tweets           & 6,857  & 7,061  & 14,228 & 14,261 \\
		Positive Tweets          & 6,173  & 5,464  & 571    & 488    \\
		Negative Tweets          & 1,970  & 2,475  & 201    & 251    \\ \hline
		Uppercased Tokens Total & 38,519            & 32,467            & 36,388            & 30,177            \\
		Uppercased Phrases Total & 861    & 1,019  & 406    & 953    \\ \hline
		\#URL\# Token            & 16,565 & 17,018 & 10,887 & 13,900 \\
		\#HASHTAG\# Token        & 6,739  & 4,715  & 5,905  & 1,580  \\
		\#USER\# Token           & 5,628  & 2,279  & 10,668 & 5,949  \\ \hline
		Retweets (RT)            & 2,383  & 1,158  & 4,289  & 1,977  \\ \hline
		NER ORG            & 8,340   & 7,299  & 2,617  & 2,595  \\ 
		NER PERSON            & 7,742   & 9,801   & 4,845  & 5,573  \\ 
		NER LOC            & 188   & 222  & 5,337  & 5,214  \\ \hline
	\end{tabular}
\end{table}

\noindent The observations of the corpus content were the following: 
\begin{itemize}
	\item Fake news spreaders:
	\begin{itemize}
		\item mention other Twitter users less often (\emph{\#USER\#}\footnote{e.g. ``@Username''}).
		\item utilize fewer hashtags (\#HASHTAG\#).
		\item re-post fewer tweets (RT).
		\item share slightly more URLs (\#URL\#).
	\end{itemize}
	\item Spanish speaking authors use more emojis than English speaking Twitter users.
	\item Half of the English tweets are based on factual information and most of the Spanish tweets (90\%) are free of emotions.
	\item Fake news tend to be more often negative. 
	\item Tweets of true news spreaders tend to be more often positive.
	\item By counting the named entities no significant difference between the classes could be established. 
	\item Fake news spreaders tend to tweet slightly more often about other people.  
	\item Uppercased tokens are shared equally by true news and fake news spreaders.
	\item Spanish fake news spreaders make more often use of capitalized phrases.      
\end{itemize}

 %3
\section{Preprocessing and Feature Extraction}
\label{Preprocessing}
\noindent The preprocessing pipeline was performed for both languages (EN and ES) basically. The steps for cleaning and structuring the data were performed as follows:  

\begin{enumerate}
	\item First, we extracted the text from the original XML document of each user and concatenated all 100 tweets to a single text. 
	\item White space between tokens were normalized to a single space.  
	\item URLs, hashtags and user mentions were left untouched as they are already replaced by placeholders by default. 
	\item Numbers and emojis were replaced by the placeholders \#NUMBER\# and \#EMOJI\#.
	\item Irrelevant signs, e.g. \textit{``+,*,\//,''} were deleted.
	\item Sequences of repeated characters with a length greater than three were normalized to a maximum of two letters (e.g. ``LOOOOOOOOL'' to ``LOOL'').
	\item Words with less than three characters were ignored.
	\item Stopwords were deleted by using the NLTK (Natural Language Toolkit) library\footnote{\url{https://www.nltk.org/}} for each language separately. 
	\item From the NLTK library we additionally used the \textit{TwitterTokenizer} to tokenize the words. The tokenizer is suitable for Twitter and other casual speech that is often used in social networks. Additionally, \textit{TwitterTokenizer} contains different regularization and normalization features. We made use of the lowercaser.
\end{enumerate}

\noindent After the Twitter texts were preprocessed, we tested different vectorization techniques with manual hyperparameter tuning, and by employing scikit-learn's grid search function. The hyperparameters were tuned separately for English and Spanish, but the features we used were mainly language-independent which means that the same set of features can be used in multi-language domains. The selected features were presented in Section \ref{Dataset} (e.g. counts of tokens or named entities). The only language dependant feature we experimented with was the sentiment polarity calculated separately for every tweet (whether it is positive, negative, or neutral). Besides the handcrafted features, we also experimented with automatically learned features i.e. term frequency distribution (tf) and character and word $n$-grams. Additionally, we made use of \emph{Feature Union}\footnote{\url{https://scikit-learn.org/stable/modules/generated/sklearn.pipeline.FeatureUnion.html}} to experiment with feature concatenation. To convert the tokens to a numerical matrix in order to build a vector for each language, we made use of:   

\begin{itemize}
	\item[(1)] Scikit-learn’s term frequency-inverse document frequency (TF-IDF) 
	\item[(2)] \emph{GloVe}\footnote{https://nlp.stanford.edu/projects/glove} (Global Vectors for Word Representation) word vectors pre-trained on Twitter data as well as custom trained  \emph{word2vec}\footnote{https://radimrehurek.com/gensim/models/word2vec.html} word embeddings
	\item[(3)] Scikit-learn’s Count Vectorizer   
\end{itemize}   

\noindent All tested features and their representations are summarized in Table \ref{tab:Feature Representation}. 

\begin{table}[h!]
	\centering
	
	\caption{Features, vectorization techniques and model hyperparameters used for training purposes}
	\label{tab:Feature Representation}
	
	\begin{tabular}{p{3cm}p{3cm}p{5cm}}
		\hline
		\textbf{Features} & \textbf{Vectorizer} & \textbf{Hyperparameters / ranges}              \\ \hline
		Tokens            & Word Embeddings         & n-gram\_range: $[1;3]$,$[2;7]$,$[3;7]$		\\ \hline
		Token $n$-grams     & TF-IDF   & min\_df: 1,2,3                \\ \hline
		Character $n$-grams & Count Vectorizer  & max\_features: $[1,000;50,000]$ \\ \hline
		
	\end{tabular}
\end{table}

 %4 
\section{Methodology}
\label{Methodology}
\noindent We defined the author profiling task as a binary problem predicting whether a tweet was composed by a fake news spreader or a reliable Twitter user. For each language (EN and ES) a separate classification model was trained. As mentioned before, for training and testing, we split the data in the ratio 70/30. We tested different features, vectorization techniques and dimensionality sizes in combination with a Support Vector Machine (SVM) and Logistic Regression of which we report the best performed ones. For the final SVM, we used a linear kernel with default hyperparameter values\footnote{https://scikit-learn.org/stable/modules/generated/sklearn.svm.\\LinearSVC.html}. Logistic Regression was also trained by utilizing default hyperparameters\footnote{https://scikit-learn.org/stable/modules/generated/sklearn.linear\_model.\\LogisticRegression.html}. 

The performance of the fake news spreader author profiling task was ranked by accuracy. Table \ref{tab:endresults} shows the scores for our final system performed on the official PAN 2020 test set on the TIRA platform \cite{tira2019}. Accuracy scores were calculated individually for each language by discriminating between the two classes. Each model was trained on 70\% of the training data. Hyperparameters were tuned on the remaining 30\% split. As the data set is hidden, the four confusion matrix values (TP, TN, FP and FN) and other metrics like Precision and Recall cannot be provided. Therefore, we display these classification results and accuracy scores which we achieved on the 30\% test dataset (see Table \ref{tab:testset}). The highest accuracy in English was obtained using SVM with TF-IDF weighted character $n$-grams with range $[1;3]$ and top 3,000 features. In Spanish, the best results were achieved using Logistic Regression employing a feature union of TF-IDF weighted character $n$-grams with range $[1;3]$ and top 5,000 features and a vector consisting of character $n$-gram counts with range $[3;7]$ and top 50,000 features. The submitted models achieve an overall accuracy of 73\% and 79\% on the English and Spanish corpus, respectively.

\begin{table}[]
	\centering
	\caption{Accuracy (Acc.) scores of the final submitted systems on the official PAN 2020 test dataset on Tira}
	\label{tab:endresults}
	\begin{tabular}{clcl}
		Model               & \multicolumn{1}{c}{Features}                                                                                                                                 & \multicolumn{1}{l}{Language} & \multicolumn{1}{c}{Acc.} \\ \hline
		SVM                 & TF-IDF char $n$-grams {[}1;3{]} 3,000 features                                                                                                               & EN                           & \textbf{0.73}            \\ \hline
		Logistic Regression & \begin{tabular}[c]{@{}l@{}}Feature union TF-IDF char $n$-grams {[}1;3{]}\\ 5,000 features and \\ char $n$-gram counts {[}3;7{]} 50,000 features\end{tabular} & ES                           & \textbf{0.79}            \\ \hline
	\end{tabular}
\end{table}

\begin{table}[]
	\centering
	\caption{Evaluation results on the test split of the submitted systems for every language (EN and ES) with the metrics Precision (P), Recall (R), Accuracy (Acc.) and $F_{1}$-Score}
	\label{tab:testset}
	\resizebox{\textwidth}{!}{%
		\begin{tabular}{cllllllllll}
			\hline
			\multicolumn{1}{l}{} &                                                                                                                                                              &                        & \multicolumn{4}{l}{\textbf{Confusion Matrix}}                                                     &                       &                       &                        &                          \\ \cline{4-7}
			Model                & \multicolumn{1}{c}{Features}                                                                                                                                 & Language               & \multicolumn{1}{c}{TP} & \multicolumn{1}{c}{TN} & \multicolumn{1}{c}{FP} & \multicolumn{1}{c}{FN} & \multicolumn{1}{c}{P} & \multicolumn{1}{c}{R} & \multicolumn{1}{c}{$F_{1}$} & \multicolumn{1}{c}{Acc.} \\ \hline
			SVM                  & TF-IDF char $n$-grams {[}1;3{]} 3,000 features                                                                                                               & \multicolumn{1}{c}{EN} & 35                     & 35                     & 10                     & 10                     & 0.78                  & 0.78                  & 0.78                   & \textbf{0.78}            \\ \hline
			Logistic Regression  & \begin{tabular}[c]{@{}l@{}}Feature union TF-IDF char $n$-grams {[}1;3{]}\\ 5,000 features and \\ char $n$-gram counts {[}3;7{]} 50,000 features\end{tabular} & \multicolumn{1}{c}{ES} & 42                     & 36                     & 9                      & 3                      & 0.92                  & 0.80                  & 0.86                   & \textbf{0.87}            \\ \hline
		\end{tabular}%
	}
\end{table}
  
 %5 
\include{texfiles/Other_Methods} %7 
\include{texfiles/Discussion} %8
\include{texfiles/Acknowledgements}
\endgroup
%%%%%%%%%%%%%%%%%%%%%%%%%%%%%%%%%%%%%%%%%%%%%%%%%%%%%%%%%%%%%%%%%%%%%%%%%%

\bibliographystyle{splncs03}
\begin{raggedright}
\bibliography{references}
\end{raggedright}

\end{document}